\documentclass[10pt, conference, compsocconf]{IEEEtran}

\usepackage{cite}
\usepackage{amsmath,amssymb,amsfonts}
\usepackage{algorithmic}
\usepackage{graphicx}
\usepackage{textcomp}
\usepackage{xcolor}

\usepackage{booktabs} 
\usepackage{hyperref}
\usepackage{url}
\usepackage{adjustbox}
\usepackage{makecell}

\begin{document}
%
\title{IL-Net: Using Expert Knowledge to Guide the Design of Furcated Neural Networks}


\author{{\large \bf Khushmeen Sakloth,\textsuperscript{1}
Wesley Beckner,\textsuperscript{1}
Jim Pfaendtner,\textsuperscript{1,2}
Garrett B. Goh\textsuperscript{2,*}}\\
{\small \textsuperscript{1} University of Washington (UW), \textsuperscript{2} Pacific Northwest National Lab (PNNL)}\\
{\small ksakloth@gmail.com, wesleybeckner@gmail.com, jpfaendt@uw.edu, \underline{garrett.goh@pnnl.gov}}
}

\maketitle

\begin{abstract}

Deep neural networks (DNN) excel at extracting patterns. Through representation learning and automated feature engineering on large datasets, such models have been highly successful in computer vision and natural language applications. Designing optimal network architectures from a principled or rational approach however has been less than successful, with the best successful approaches utilizing an additional machine learning algorithm to tune the network hyperparameters. However, in many technical fields, there exist established domain knowledge and understanding about the subject matter. In this work, we develop a novel furcated neural network architecture that utilizes domain knowledge as high-level design principles of the network. We demonstrate proof-of-concept by developing IL-Net, a furcated network for predicting the properties of ionic liquids, which is a class of complex multi-chemicals entities. Compared to existing state-of-the-art approaches, we show that furcated networks can improve model accuracy by approximately \textasciitilde20-35\%, without using additional labeled data. Lastly, we distill two key design principles for furcated networks that can be adapted to other domains.
\end{abstract}


%
\IEEEpeerreviewmaketitle

\section{Introduction}

Despite decades of research, designing chemicals with specific properties or characteristics is still heavily driven by serendipity and chemical intuition. Recent years have seen a resurgence in the application of machine learning to the discovery and design of new chemicals and materials. Efforts such as Materials Genome Initiative (MGI) \cite{de2014} have led to the creation of public data repositories like the Harvard Clean Energy Project,\cite{hachmann2011} and Open Quantum Materials Database.\cite{saal2013} These repositories comprise of material properties that can be optimized during a design process.

\subsection{On Chemical Complexity}

However, most of the large data repositories are of solid crystalline materials or single molecules, and many important chemicals and materials for a wide range of applications are liquids. Liquids are multi-chemical entities, and are significantly more complex to model, as they comprise of different independent chemicals that interact with one another to give rise to a collective behavior or property of the liquid. One example of industrial relevance is ionic liquids (ILs). Popularly termed as ‘green solvents’ due to their exceptional properties (e.g. negligible vapor pressure, high chemical stability), ILs comprise of charged chemicals(paired ions) that are liquid at or above room temperature. In contrast to solid materials, because of their complex and dynamic nature, ILs present a host of challenges that prevent direct mimic of the successful MGI-type approaches.

\subsection{Industrial Applications of Ionic Liquids}

Ionic liquids (ILs) have wide applications in industry, and are used in many engineering (i.e. supercapacitors, solar thermal energy stations, \cite{ahmed2017recent}) and biotechnology (i.e. bio-catalysis/enzyme stabilization, nanomaterials synthesis, bioremediation.\cite{abo2015potential}) verticals. ILs have also shown promise in renewable energy technologies as the liquid component of redox flow batteries and concentrated solar power. \cite{chakrabarti2014prospects,paul2017enhanced}. Particularly for clean energy technlogies, the thermodynamic (heat capacity, density) and transport (viscosity) properties of ILs are especially important, as it has a direct impact on device efficiency.\cite{wang2013recent}.

To accelerate the property predictions for ILs, various equation of state models (rule-based models) were developed.\cite{maia2012} These models were highly accurate but could not be generalized across different IL designs. Therefore, the use of more sophisticated simulation methods to predict these properties is crucial for realizing IL potential in these applications: there are theoretically $10^{14-18}$ possible IL structures which makes it unfeasible to iterate through using wet-lab experiments.\cite{karunanithi2013computer,tian2012exploring}

In some cases, physics-based simulations such as molecular dynamics (MD) and monte carlo (MC) simulations can supplant wet-lab measurements. MD/MC have been shown to be highly accurate in the calculation of some properties such as density.\cite{sprenger2015general}, Howevever, MD/MC in many cases, fall short in handling properties such as viscosity that are calculated from fluctuation-dissipation theorem \cite{zhang2015reliable}. Therefore, while both rule-based models and physics-based simulations have met with some success in predicting IL properties, the use of DNN models will provide an added advantage.

\subsection{Machine \& Deep Learning in Chemistry}
For the prediction of chemical properties that cannot be easily computed through physics-based or rule-based methods, machine learning (ML) methods have been used to correlate structural features with the activity or property of the chemical. This approach is formally known as  Quantitative Structure-Activity or Structure-Property Relationship (QSAR/QSPR) modeling in the chemistry literature~\cite{cherkasov2014}.
Molecular descriptors are engineered features developed based on first-principles knowledge, which typically are basic computable properties or descriptions of a chemical's structure, and they are used as input in QSAR/QSPR models. At present, over 5000 molecular descriptors have been developed ~\cite{todeschini2008}, and other features such as molecular fingerprints have also been designed and used in training ML models.~\cite{rogers2010}. More recently, DNN models have also been developed~\cite{dahl2014, ramsundar2015, hughes2016}. On average, DNN models either perform at parity or slightly outperform prior state-of-the-art ML models~\cite{goh2017r}. More recent research has focused on leveraging representation learning from raw data instead of relying on molecular descriptors, representing chemicals as graphs~\cite{duvenaud2015, kearnes2016}, images~\cite{goh2017c1, goh2017c2, wallach2015}, or text~\cite{jastrzkebski2016, goh2017s}. Other developments also include advanced learning methods like weak supervised transfer learning~\cite{goh2017c3} and multimodal learning. 

Despite the rapid advancements in this field, the majority of the literature has been focused on property prediction of simple chemicals, like solid crystalline matter and simple molecules.

\subsection{Contributions}

Our work improves the existing state of modeling ILs properties by using expert knowledge in guiding the design of the DNN architecture. \textit{Specifically, guided by our knowledge of chemical interactions, we developed a furcated network architecture that is more accurate than current state-of-the-art network architectures and modeling approaches}. Our specific contributions are as follows. 
\begin{itemize}
	\item We develop the first furcated neural network architecture for modeling complex multi-chemical entities, such as ionic liquids (ILs).
	\item We investigate the effect of network architecture design and hyperparameters on model accuracy, and demonstrate that furcated DNN models outperforms conventional DNN models by up to 35\%.
	\item We develop a novel overweighted multi-task learning approach for maximzing model accuracy, which consistently outperforms the baseline model by 20-35\%.
\end{itemize}

The organization for the rest of the paper is as follows. In section 2, we outline the motivations and design principles behind developing a furcated neural network that incorporates considerations from expert knowledge. Next, we examine the ILThermo dataset used for this work, its applicability to chemical-affiliated industries, as well as the training methods used. Lastly, in section 3, using IL property prediction as an example, we explore the effectiveness of furcated network designs, and develop overweightedmulti-task learning approach to improve model accuracy. We conclude with the development of IL-Net, and evaluate its performance against the current state-of-the-art approaches in the field.

\subsection{Related Work}

Most of the existing literature on using neural networks to model complex multi-chemical entities like ILs has been limited to a single property, and typically has a narrow scope in terms of applicable chemicals. For example, viscosity prediction have been reported in the literature,\cite{matsuda2007computer,gardas2009group}. While many of these earlier models are still considered to be one of the most accurate temperature dependent viscosity models to date, there is typically a narrowly defined set of ILs for which the models is applicable.\cite{gardas2009group} Therefore, more recent efforts have focused on expanding the scope of these models.\cite{beckner2018statistical}. It should also be emphasized that all prior literature have not looked into neural network architecture design and are using the standard feedforward MLP network design, varying only hyperparameters like number of layers and depth of the network.

Recent publications have also indicated that multi-task models for predicting chemical properties can confer accuracy improvements. \cite{dahl2014, ramsundar2015, hughes2016} Namely, this performance improvement seems to depend on whether the outputs are correlated, so that the neural network can borrow ‘signal’ from molecular structures in the training data of the other outputs.\cite{xu2017demystifying} In the context of IL modeling, some examples include developing a ‘global model’ for a specific class of chemicals for predicting viscosity, density and thermal expansion coefficient, \cite{cancilla2015} or for predicting density, viscosity and electrical conductivity.\cite{kianfar2018} Multi-task modeling of IL properties (density, viscosity and refractive index) have also seen mixed success compared to single-task models.\cite{dopazo2014}

\section{Methods}

In this section, we provide details about the design principles behind the furcated neural network design. Then, we document the training methods, as well as the evaluation metrics used in this work.

\subsection{Extracting Design Principles from Domain Knowledge}
In the absence of additional data, network architecture plays a key role in improving the accuracy of various DNN models. One example is in the annual ILSVRC assessment, where using the same ImageNet dataset, network architecture advances was responsible for improving the state-of-the-art in image recognition to the point beyond human-level accuracy. Advances like GoogleNet (Inception architecture)~\cite{szegedy2015}, ResNet (Residual link architecture)~\cite{he2015} and DenseNet~\cite{huang2017} are all examples on how one can develop network architecture to improve model accuracy. Despite these advances, network design remains more of an art than a science.

In our work, we use expert knowledge about chemical interactions of ILs to guide us in the high-level network design decisions. The hypothesis behind this approach stems from the fact that neural network develop hierarchical representations. This can be interpreted as a flow of information and concepts, starting from the more basic representations in the lower layers that help develop more advanced representations in the upper layers. This process of learning hierarchical representation may also mirror how human experts learn. In established chemistry research, there is also a similar ordering of concepts, specifically the hierarchical nature of chemical structures, from atoms to molecules, and also in chemical interactions, from intramolecular to intermolecular interactions.

Therefore, we can explore the hierarchical ordering of chemistry knowledge and use it to design a similarly ordered furcated neural network. We begin by examining various stages in constructing chemical structures, as there is usually a correlation between structure and property. Starting at the lowest stage, the atom, which is the basic unit in chemistry, one can construct a collection of atoms that operate as an independent unit. These independent units are known as cations, anions or molecules in the literature. At the next stage, units are also the constituent components of ILs, and ILs typically have 2 or more different units. There is also a similar parallel in chemical interactions, which from domain knowledge we know is key in understanding and predicting IL properties. At the lowest stage, atoms in each unit interact with one another and collectively govern the behavior of the unit. This is known as intramolecular interaction in the literature. At the next higher stage, different units of the IL interact with one another, and collectively govern the properties of the IL. This is known as intermolecular interaction in the literature.

Based on these fundamental understandings of chemical structures and interactions, we will map each stage of chemical structure/interaction to a sub-network. Here, sub-network refers to a specific portion of the deep neural network which is ‘designated’ to learn representations at a specific level of chemical hierarchy, and it only receives input data pertaining to hierarchy. Then, the sub-networks are arranged in a specific order to control the flow of information and learnable representations. In the context of our work, the resulting network designs are bifurcated, hence the term Furcated Neural Networks.

\subsection{Designing the Furcated Neural Network}
\begin{figure}[!htbp]
\centering
\includegraphics[scale=0.6]{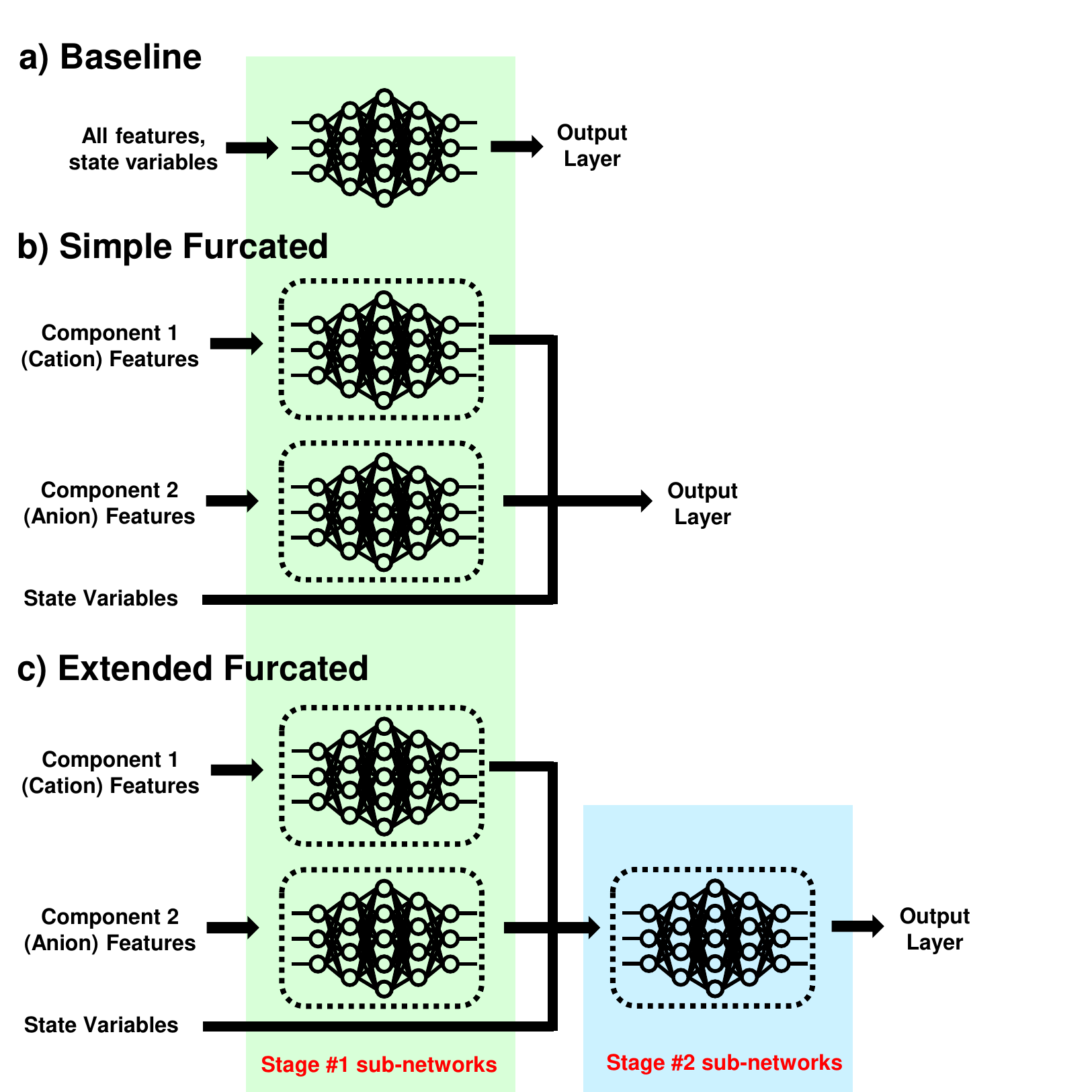}
\caption{\small Schematic diagram of standard feedforward MLP networks used as (a) baseline model, furcated networks explored in this work of (b) simple and (c) extended design.}
\label{fig:1}
\end{figure}

The current approach of using standard feedforward MLP networks, as illustrated in Figure ~\ref{fig:1} forms the baseline model of our work. For modeling IL properties, engineered features (molecular descriptors) associated with all the chemical components of ILs, along with the state variables (i.e. non-chemical parameters like pressure and temperature) are collectively used as the input data.

Next, based on the design principles elaborated on earlier, we designate sub-networks for each component of the ILs. The ILs modelled in this work have two component (cations and anions), and we have a separate sub-network for each. Each sub-network also only receives input data corresponding to the features for their designated component, and because there is no mixing of information between different components of the ILs, it can only learn representations pertaining to itself (intramolecular interactions). Next, the output of these stage \#1 sub-networks are passed on to another stage \#2 sub-network. In stage \#2, the sub-network receives representations from both components and therefore learn representations that govern the behavior between components (intermolecular interactions). In addition, the stage \#2 sub-network also receives input data about the state-variables that are passed in as separate ‘3rd-rail’. This decision to inject state-variables information at stage \#2 was intentional, as from domain knowledge we know that state-variables change the dynamics and hence interactions between components, but have little effect in changing the independent behavior of each component. The resulting network design, hereon referred to as the Extended Furcated model is illustrated in Figure ~\ref{fig:1}.

From domain knowledge, we know that modeling the behavior between components is usually important for predicting IL properties, and non-linear interactions are usually more challenging to model. Therefore, we created a variant of the Extended Furcated model by removing the stage \#2 sub-network to create the Simple Furcated model. This network will be unable to learn sophisticated representations about the interactions between the components of the ILs, although it can still perform a linear combination using its final output layer. It is also expected that since IL properties are highly dependent on interactions between their constitutent components, the inability to model this process explicitly should lead to a noticable decline in model accuracy.

In terms of model hyperparameters, for each network or sub-network, we performed a grid search for (2,3,4,5) fully-connected layers, and (16,32,64,128,256,512) neurons per layer with ReLU activation functions. A dropout of 0.5 was added after each layer to mitigate overfitting. The results reported in this paper correspond to the best set of model hyperparameters identified, which is defined as having the lowest validation loss.

\subsection{Dataset Preparation}
The open source ILThermo database is a web-based database cataloging IL properties, and it is maintained by the National Institute of Standards and Technology (NIST).~\cite{dong2007} We compiled a curated ILThermo dataset by web scraping for IL properties that have substantial ($>$10,000) data points. The final curated dataset (see Table~\ref{table:1}) has 23,000 entries at different physical conditions (temparature, pressure) for 3 different IL properties (heat capacity, density, viscosity).

\begin{table}[!t] 
		\begin{center}
		\begin{tabular}{|c|c|c|c|}
				\hline
				Condition & Min & Max & Size \\
				\hline\hline
				Temperature & 278.15 K & 373.15K & 23,982 \\
				\hline
				Pressure & 100 kPa & 20000 kPa & 23,982 \\
				\hline\hline
				Property & Min & Max & Size \\
				\hline\hline
				Heat Capacity & 231.8 J/K.mol & 1764 J/K.mol & 23,982 \\
				\hline
				Density & 847.5 kg/m\textsuperscript{3} & 1557.1 kg/m\textsuperscript{3} & 23,982  \\
				\hline
				Visocity & 0.00316 Pa/s & 10.2Pa/s & 23,982  \\
				\hline
		\end{tabular} 
		\end{center}
\caption{Characteristics of the curated ILThermo dataset used in this work.}
\label{table:1}
\end{table}

\subsection{Feature Calculation}
Saltyt~\cite{salty2018}, an interactive data exploration tool for ILs data, built on top of RDKit~\cite{landrum2016}, an open-source cheminformatics software was used to generate the input features. Using salty, we computed 94 engineered features (molecular descriptors) for each component (cation and anion) of each IL entry, and included a broad set of features from all major descriptor classes. In addition, the features were scaled to have a mean of zero and unit variance before it is used for training. Lastly, the labels to be predicted (heat capacity, density, viscosity) were also log transformed. 


\subsection{Training the Neural Network}

From the original curated dataset, 25\% was separated to form the test set and the remaining 75\% was used for training and validation. We used a random 5-fold cross validation approach for training and evaluated the performance of the model using the validation set. 

Our models were trained using a Tensorflow backend~\cite{abadi2016} with GPU acceleration. The network was created and executed using the Keras 1.2 functional API interface~\cite{chollet2015}. We use the Adam algorithm~\cite{kingma2014} to train for 500 epochs with a batch size of 30, using the standard settings recommended: learning rate = 10\textsuperscript{-3}, $\beta$ = 0.9, $\beta$ = 0.999. We also included an early stopping protocol to reduce overfitting. This was done by monitoring the loss of the validation set, and if there was no improvement in the validation loss after 50 epochs, the last best model was saved as the final model.

The mean squared error loss function was used for training. In typical multi-task learning, there is equal weights applied to the loss function for each task. However, we observed that the typical losses attained during single-task learning across all 3 tasks vary by at least an order of magnitude. In order to ensure balance between the tasks, we used a reweighted loss function as defined below:

Loss = $\frac{y_{pred}^2 - y_{exp}^2}{N_1}$ + $\frac{y_{pred}^2 - y_{exp}^2}{N_2}$ + $\frac{y_{pred}^2 - y_{exp}^2}{N_3}$

The metric used to evaluate the performance between different models is root-mean-squared-error (RMSE).


\section{Results and Discussion}

In this section, we conduct several experiments to determine the hyperparameters for the best model of each class (baseline, simple, extended). Once identified, we compare the furcated network performance across the 3 tasks (heat capacity, density, viscosity). Then, we evaluate the effectiveness of multi-task learning, and develop a novel overweighted multi-task learning approach for maximizing model accuracy.

\subsection{Model Architecture Exploration}

Using the range of hyperparameters listed in the previous section, Table~\ref{table:2} summarizes the hyperparameters for the best model for each model class. In our initial single-task learning models, separate hyperparameter searches were performed for each task, and so a total of 9 best models were identified, with 3 model classes for each of the 3  tasks. In our latter multi-task learning models, hyperparameter searches were performed singularly across all tasks, resulting in 1 best model.


\begin{table*}[t] 
		\begin{center}
		\begin{tabular}{|c|c|c|c|c|c|c|c|}
				\hline
				Property & Model & Stage \#1 & Stage \#2 & Hyperparam Optimization & Val RMSE & Test RMSE & \% Improvement \\
				\hline\hline
				Heat Capacity & Baseline & 2(64) & - & Best for each task & 0.058 & 0.062 & -\\
				Heat Capacity & Simple & 2(64) & - & Best for each task & 0.060 & 0.061 & 1\% \\
				Heat Capacity & Extended & 2(64) & 3(64) & Best for each task & 0.045 & 0.045 &28\% \\
				Heat Capacity & Extended-ST & 2(64) & 2(128) & Best for all tasks & 0.047 & 0.048 & 23\% \\
				Heat Capacity & Extended-MT & 2(64) & 2(128) & Best for all tasks & 0.048 & 0.048 & 22\% \\
				Heat Capacity & Extended-MT-OW & 2(64) & 2(128) & Best for all tasks & 0.042 & 0.042 & 33\% \\
				\hline
				Density & Baseline & 2(64) & - & Best for each task & 0.0130 & 0.0130 & - \\
				Density & Simple & 2(128) & - & Best for each task & 0.0190 & 0.0189 & -45\% \\
				Density & Extended & 2(128) & 3(32) & Best for each task & 0.0130 & 0.0130 & 0.2\% \\
				Density & Extended-ST & 2(64) & 2(128) & Best for all tasks & 0.0149 & 0.0152 & -17\% \\
				Density & Extended-MT & 2(64) & 2(128) & Best for all tasks & 0.0129 & 0.0135 & -4\% \\
				Density & Extended-MT-OW & 2(64) & 2(128) & Best for all tasks & 0.0126 & 0.0128 & 1.6\% \\
				\hline
				Viscosity & Baseline & 2(128) & - & Best for each task & 0.156 & 0.160 & - \\
				Viscosity & Simple & 2(64) & - & Best for each task & 0.221 & 0.218 & -41\% \\
				Viscosity & Extended & 2(64) & 2(128) & Best for each task & 0.129 & 0.133 & 17\% \\
				Viscosity & Extended-ST & 2(64) & 2(128) & Best for all tasks & 0.130 & 0.134 & 16\% \\
				Viscosity & Extended-MT & 2(64) & 2(128) & Best for all tasks & 0.142 & 0.147 & 9\% \\
				Viscosity & Extended-MT-OW & 2(64) & 2(128) & Best for all tasks & 0.124 & 0.127 & 21\% \\
				\hline
		\end{tabular} 
		\end{center}
\caption{Summary of the best models attained. Hyperparameters of various models, for each stage (see Figure ~\ref{fig:1} for schematics) specifying the number of layers and number of neurons per layer, for example 3(128) denotes 3 layers of 128 neurons per layer. Different training methods were also used, ST is single-task, MT is multi-task and OW is overweighted.}
\label{table:2}
\end{table*}

\begin{figure}[!htbp]
\centering
\includegraphics[scale=0.4]{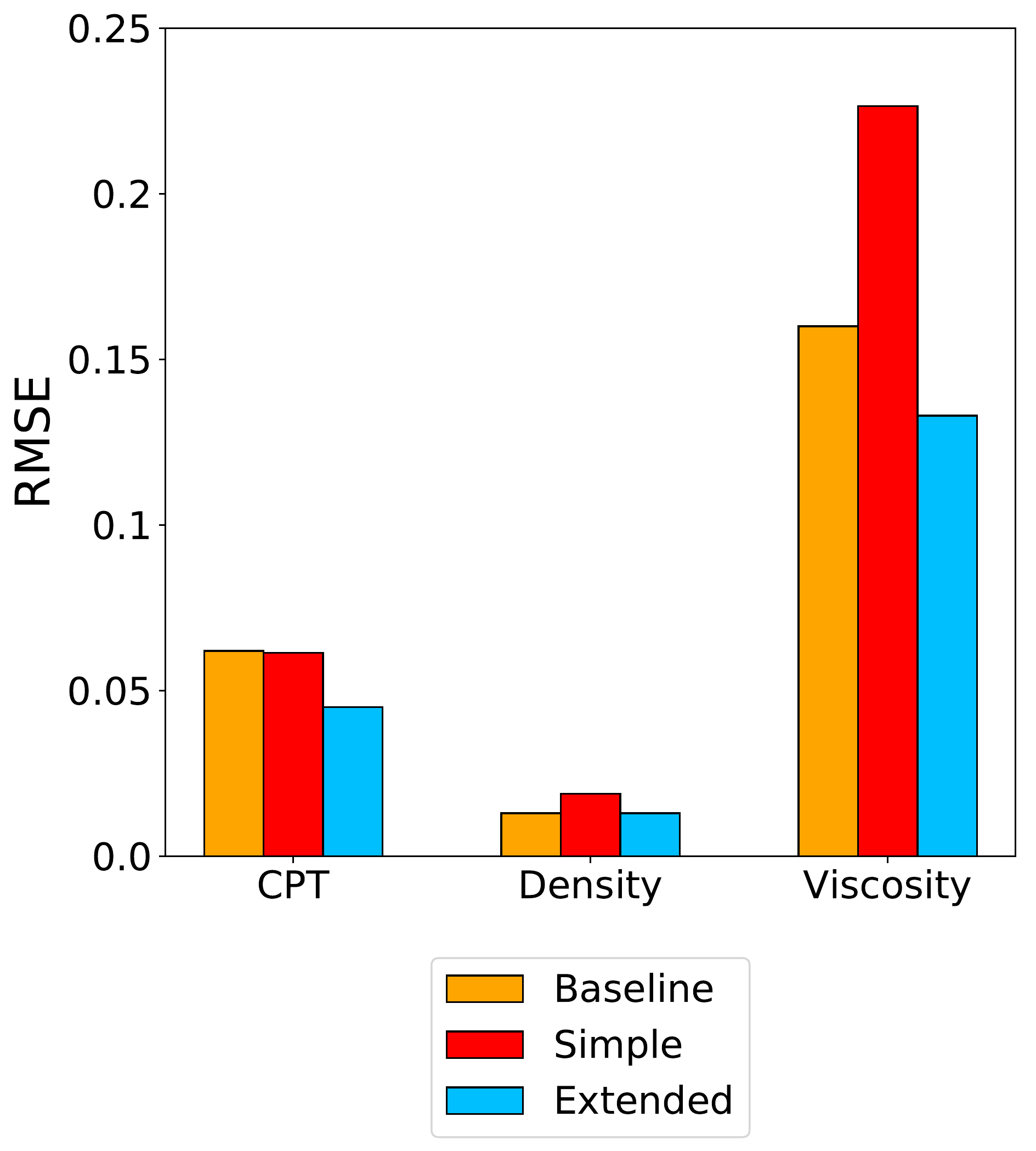}
\caption{\small Validation and test RMSE for baseline, simple and extended furcated single-task models evaluated for heat capacity (CPT), density and viscosity predictions.}
\label{fig:2}
\end{figure}

From results as summarized in Figure ~\ref{fig:2}, we observed that the validation RMSE and test RMSE is comparable for all 9 models evaluated, indicating that there was no signficant overfitting. We observed that the simple furcated model significantly underpeforms the baseline model for density and viscosity predictions, with a RMSE that is typically 40\% higher. On the other hand, the extended furcated model outperforms the baseline model. For heat capacity, the extended model achieved a val/test RMSE of 0.045/0.045 J/Kmol against the baseline value of 0.058/0.062 J/Kmol. Density results were comparable to baseline, with the extended model attaining a val/test RMSE of 0.0130/0.130 kg/m\textsuperscript{3} against the baseline value of 0.0130/0.0130 kg/m\textsuperscript{3}. For viscosity, the val/test RMSE of 0.129/0.133 Pa/s is better than the baseline value of 0.156/0.160 Pa/s. Across all tasks, we observed that the extended furcated model is more accurate by about \textasciitilde20-30\% compared to the baseline model, but in some cases like density predictions there is no improvement.

As mentioned in preceding section, the properties of ILs are closely tied to the interactions between its constituent components. Therefore, the improved accuracy as one moves from simple furcated to baseline to extended furcated model is consistent with expectations from domain knowledge. 

The key architectural difference between the simple and furcated model is the addition of the stage \#2 sub-network to specifically model interactions between components. The trends in model accuracy proves that without the ability to model these interactions, accuracy decreases signficantly. In comparison to the baseline model, the input features of both components are fed into the same network and thus representations to model interactions between components can be learned. However, this may not be straightforward as the network will need to implicitly learn how to classify input features based on components, before it can develop representation to explain the interactions different components.

This is different in the extended furcated model, the features are explicitly segregated by component, with designated sub-networks to learn representations. Therefore, by controlling the flow of data and learnable representations by using the extended furcated network design, our results demonstrate that model accuracy can be improved without any additional labeled data.

\subsection{On Multi-task Learning}

Multi-task learning, which trains a network simultaneously on more than one task is an established method that can improve overall model accuracy, and it works by exploiting commonalities and differences between a set of related tasks.~\cite{ruder2017} As before, hyperparameter searches were conducted by jointly optimizing across all tasks, and the best models were identified (see Table~\ref{table:2}). In training the multi-task models, the loss function also needs to be reweighted to account for the difference in loss magnitude for each task. This is because if an unweighted loss function is used, it will be biased towards viscosity predictions, which has the highest loss value. To ensure balanced weighting across the 3 tasks, and based on the typical loss values attained in the single-task models, we used a reweighting ratio of 5:30:1 for the loss of heat capacity, density and viscosity respectively.

\begin{figure}[!htbp]
\centering
\includegraphics[scale=0.4]{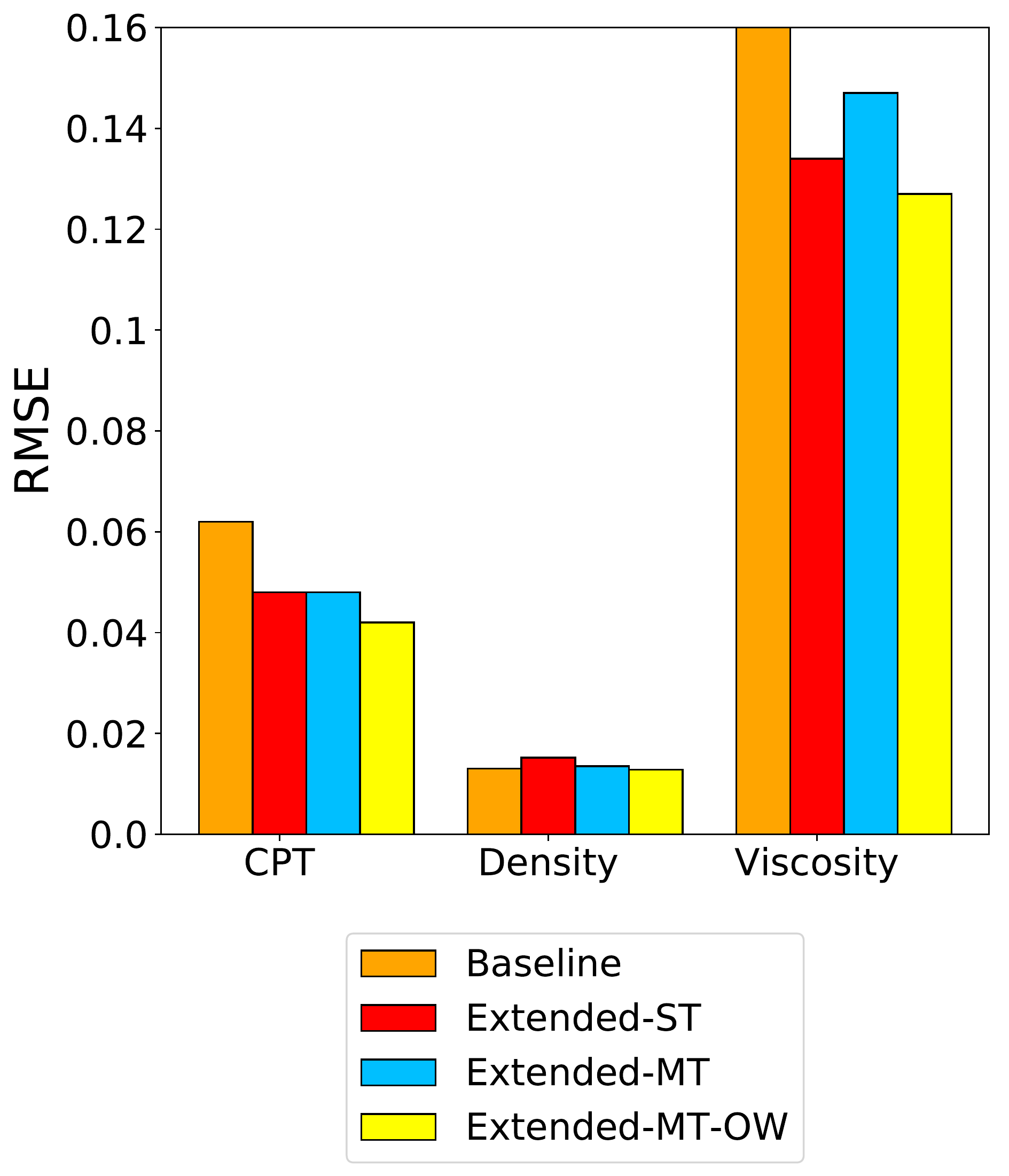}
\caption{\small Validation and test RMSE for baseline, simple and extended furcated multi-task models evaluated heat capacity (CPT), density and viscosity predictions.}
\label{fig:3}
\end{figure}

The results as shown in Figure ~\ref{fig:3} shows that the furcated network design is also effective in a multi-task setting, on the underlying assumption that the tasks to be predicted are of similar nature. Heat capacity predictions achieved a lower val/test RMSE of 0.048/0.048 J/Kmol against the baseline model of 0.058/0.062 J/Kmol. For density, the results are slightly less accurate with a val/test RMSE of 0.0129/0.0135kg/m\textsuperscript{3} against the baseline value of 0.0130/0.0130 kg/m\textsuperscript{3}. For viscosity, it achieved a lower val/test RMSE of 0.142/0.147 Pa/s against the baseline value of 0.156/0.160 Pa/s. Across all tasks, we observed that multi-task learning is more accurate by about \textasciitilde10-22\% compared to the baseline model, but this improvement may not be consistent as in the example of density where it is less accurate.

Next, we compared the results between single-task vs multi-task extended furcated models. The single-task results would not be an appropriate comparison, as the model hyperparameters may be different as they were optimized on a per task basis. Therefore, using the current multi-task network hyperparameters, we trained 3 separate models using single-task learning. By computing the percentage improvement relative to baseline (Figure ~\ref{fig:4}), we observed mixed results. While multi-task learning improved density predictions by 13\%, it was also less accurate for viscosity by 7\%. For heat capacity, the differences are negligible.

There are several factors influencing the mixed results from multi-task learning. A mitigating factor is that the ILThermo~\cite{dong2007} database curates experimental measurements from many different sources and experiments, and this may have introduced additional irreducible error in these models. In other multi-task learning models in other chemical property prediction,~\cite{ramsundar2015}, multi-task learning demonstrated consistent accuracy improvement, although a substantial quantity of labeled data and $>$40 tasks were typically required before non-trivial improvements over single-task models were achieved. Other related work in multi-task learning for IL property prediction has also reported that it underperformed single-task models.~\cite{dopazo2014} Therefore, while in principle multi-task learning should improve model accuracy, in practice this has met with mixed success. Lastly, it should be emphasized that we are using the largest publicly accessible dataset on IL properties. While acquisition of additional labeled data is a safe option to increase the effectiveness of multi-task learning, due to the infeasibility of this approach which requires substantial wet-lab experimentation, for the scope of this work this will not be explored.

\subsection{Overweighted Multi-task Learning}

We have shown that a furcated network design can provide accuracy improvements over baseline models, but its consistency seems to be dependent on the training approach (single-task vs multi-task). In order to obtain a model that is consistently better using a standardized approach, we explored the effectiveness of overweighting the loss function.

In the previous section, we applied a reweighted ratio of 5:30:1 to normalize each task, which ensures the loss from each task will be approximately the same magnitude. However, fine control over the relative magnitude of individual task losses is non-trivial due to the stochastic nature of training neural networks. Instead of attempting to control this optimization problem, we approach it with the underlying hypothesis that intentionally overweighting the loss function for a specific task will improve the accuracy of that task at the expense of others, but still retain the advantages of multi-task learning. Here, we experimented overweighting by 100X, for example, if we were to overweight a model to predict heat capacity the reweighted ratio would be modified from 5:30:1 to 500:30:1 for heat capacity, density and viscosity respectively..

Using this overweighting approach, we trained 3 separate multi-task models and the results are summarized in Figure ~\ref{fig:3}. Across all tasks, we observed a consistent improvement of approximately \textasciitilde20-35\% compared to the baseline. Specifically, heat capacity achieved the lowest val/test RMSE of 0.042/0.42 J/Kmol against the baseline model of 0.058/0.062 J/Kmol. For density, the results are slightly improved with a val/test RMSE of 0.0126/0.0128 kg/m\textsuperscript{3} against the baseline value of 0.0130/0.0130 kg/m\textsuperscript{3}. For viscosity, it achieved the lowest val/test RMSE of 0.124/0.127 Pa/s against the baseline value of 0.156/0.160 Pa/s.

Across our experiments, we also observed that density predictions were particularly challenging to improve on over the baseline model. From domain knowledge and physics-based simulations, it has been observed that density is a simpler property to calculate than heat capacity or viscosity. For example, approximate parameters used in physics-based simulations (forcefields parameters of molecular dynamics simulations) can easily attain reasonable calculations, that is not the case for heat capacity and viscosity calculations.\cite{zhang2015reliable} Thus, it is conceivable that modeling density is a simple process, from which using a more sophisticated furcated network design would provide minimal improvement. Another consideration is that the type of interactions that govern IL properties. Density is primarily determined by short-range van der Waals forces, whereas heat capacity and viscosity is primarily determined by longer-ranged electrostatic forces.\cite{chandler1983} Not only do these interactions operate at different range, the physics behind them is also different. It is plausible that the input features used in this work do not describe VDW effect as well as electrostatics, and therefore improved feature engineering to describe VDW forces may improve model performance.

\begin{figure}[!htbp]
\centering
\includegraphics[scale=0.4]{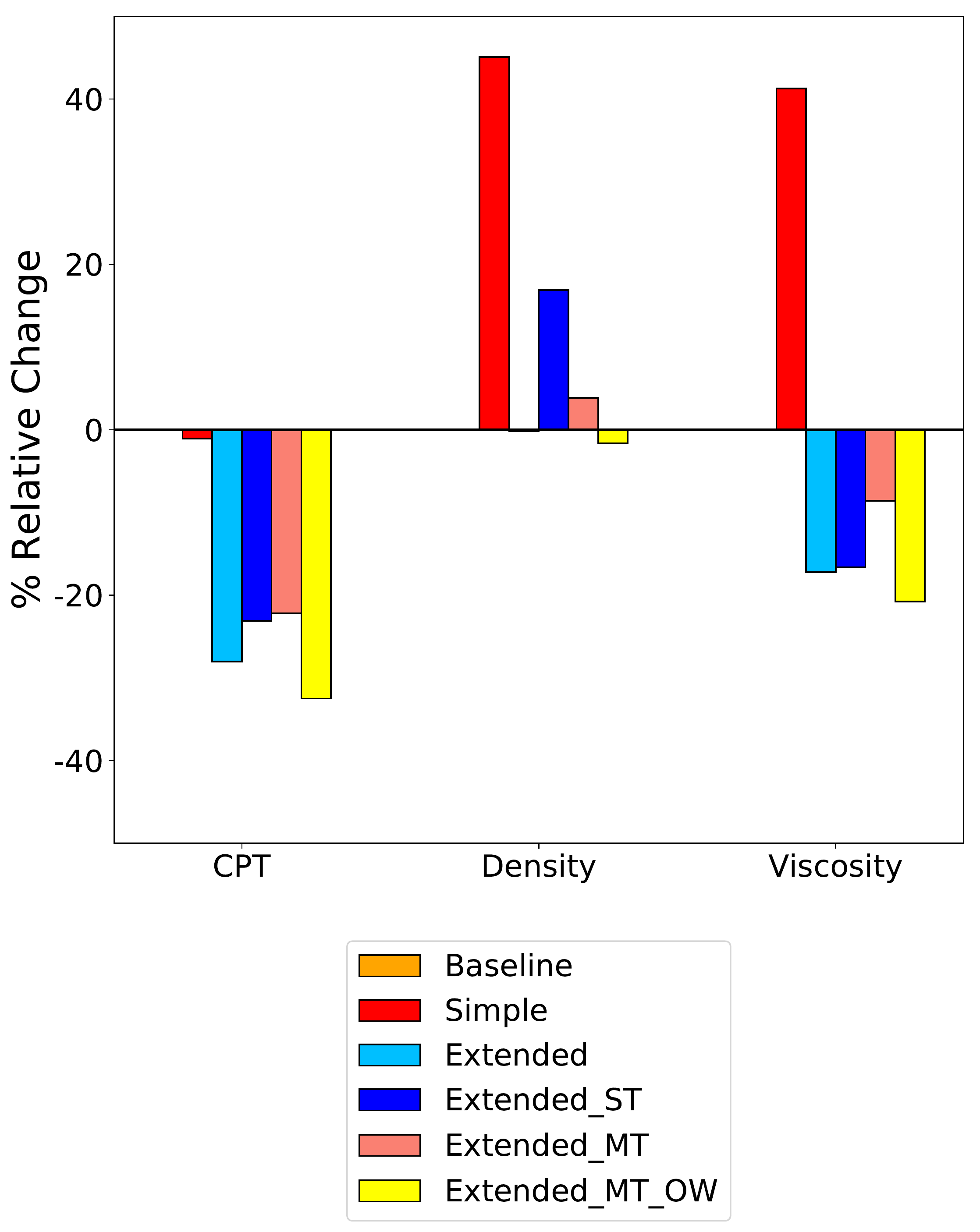}
\caption{\small Relative error improvement for all models evaluated for heat capacity (CPT), density and viscosity predictions.}
\label{fig:4}
\end{figure}

In summary, when compared to all other models evaluated in this work, our experiments as summarized in Figure ~\ref{fig:4} and Table~\ref{table:2} indicate that using an appropriate furcated network design with overweighted multi-task learning has been effective in consistently improving the accuracy of our models without requiring additional labeled data.

\subsection{Furcated Network Designs in Other Domains}

While the furcated model that we have developed is specific to IL property prediction, the design principles is applicable to other domains. Specifically, the following factors were critical in designing a successful furcated network in this work.
\begin{itemize}
	\item A good grasp from expert knowledge on the flow of data and concepts/representations is necessary. In our work, this is the hierarchy of interactions as established from chemical knowledge. In other domains, where there is a schematic or organized flow of information, such as in many engineering fields, similar principles can also be adapted to restrict the learning of representations to specific sub-networks.
	\item The ability to identify and segment data for each sub-network is also critical, as it ensures that sub-networks learn representations as designated. In our work, we computed features for the components of ILs as used separate input data for separate sub-networks. For other domain applications, we anticipate this solution will be domain-specific where combining expert-knowledge with deep learning ingenuity will be needed.
\end{itemize}

\section{Conclusion}

In conclusion, we developed a novel furcated neural network architecture, a first of its kind for use in modeling complex multi-component chemical entities. Using a subset of the ILThermo database, we illustrated our work in the context of ionic liquid (ILs) property prediction for heat capacity, density and viscosity. Combined with a novel overweighted multi-task learning approach, we demonstrated that our furcated models outperform the current approaches in the literature by approximately \textasciitilde20-35\%. Specifically, IL-Net achieved a RMSE of 0.042 JK/mol for heat capacity, 0.0128 kg/m\textsuperscript{3} for density and 0.127 Pa/s for viscosity predictions. Furthermore, we identified two key design principles for effective furcated networks. First, the hierarchy or flow of data and concepts/representations from expert knowledge is used to guide high-level architecture design choices. Second, identifying and segmenting data into specific sub-networks to control the representations it learns. By using expert-knowledge to guide neural network design, we anticipate that such furcated networks will be particularly effective in other fields where there is established high-level understanding of the subject matter.

\bibliographystyle{IEEEtran}
\bibliography{il_net}





%
%
%

\end{document}